\documentclass[letterpaper, 10 pt, conference]{ieeeconf}  %

\IEEEoverridecommandlockouts                              %

\usepackage{graphicx}
\usepackage{epsfig} %
\usepackage{amsmath} %

\usepackage{amssymb}  %
\usepackage{amsthm}
\DeclareMathAlphabet{\altmathcal}{OMS}{cmsy}{m}{n}
\usepackage{mathrsfs}
\usepackage{bm}
\usepackage{float}
\usepackage{bbm}
\usepackage{color}
\usepackage{placeins}
\usepackage{algorithm}
\usepackage[colorlinks=true,linkcolor=black,anchorcolor=black,citecolor=black,filecolor=black,menucolor=black,runcolor=black,urlcolor=cyan]{hyperref}
\usepackage{wrapfig}
\usepackage[table,xcdraw,dvipsnames]{xcolor}
\usepackage{multirow}
\usepackage{mathtools}
\usepackage{algpseudocode}
\usepackage{xspace}
\usepackage{etoolbox}
\usepackage{tikz}
\usepackage{cite}
\usetikzlibrary{calc}
\usepackage{caption}
\newtheorem{defn}{Definition}
\newtheorem{thm}[defn]{Theorem}

\renewcommand{\det}[1]{\lvert#1\rvert}

\providecommand{\R}{\ensuremath \mathbb{R}}

\newcommand{\Btwo}{B_{2}}

\newcommand{\pr}{\mathrm{P}}
\newcommand{\erf}{\mathrm{erf}}
\newcommand{\bound}{\mathcal{H}}

\newcommand{\pos}{\mathbf{p}}

\newcommand{\boldmu}{\boldsymbol{\mu}}

\usepackage{pifont}
\newcommand{\cmark}{\ding{51}}
\newcommand{\xmark}{\ding{55}}

\title{\LARGE \bf
\algname: Risk-Aware Motion Planning for Quadrotors in \\Cluttered 3D Gaussian Splats
}

\author{Seth Isaacson, William Hong, Katherine A. Skinner, and Ram Vasudevan%
\thanks{This work was supported by MCity at the University of Michigan}
\thanks{Seth Isaacson, William Hong, Katherine A. Skinner, and Ram Vasudevan are with the Department of Robotics, University of Michigan, Ann Arbor, MI 48109.}
\thanks{\{sethgi, billhong, kskin, ramv\}@umich.edu}}

\newcommand{\algname}{AirSplan\xspace}

\begin{document}

\maketitle
\thispagestyle{empty}
\pagestyle{empty}

\begin{abstract}
Quadrotors are increasingly deployed in applications such as agriculture, infrastructure inspection, and maintenance.
In each of these applications, the robot must navigate complex scene geometry while remaining strictly collision-free.
Unlike in ground domains, even minor collisions for aerial vehicles can result in the loss of the robot.
This safety requirement induces a pair of technical challenges.
First, the environment must be represented with sufficient fidelity to encode complex structure, even when no ground-truth obstacle data is available.
Second, a motion planner must leverage this representation to determine a collision-free path to the goal.
This paper proposes a system that addresses these complementary challenges.
The proposed method, \algname, adopts a normalized variant of 3D Gaussian Splatting that encodes high-fidelity scene geometry.
It then applies a novel reachability-based motion planner that leverages the differential flatness of quadrotors to compute continuous-time collision constraints that tightly overapproximate the robot's occupancy.
Experiments demonstrate that AirSplan successfully finds a path in 81.2\% of challenging test cases, a significant improvement over the nearest baseline method's 51.2\%.
\end{abstract}

\section{Introduction}

Quadrotors enable robotic operation in environments inaccessible to ground-restricted platforms, enabling applications such as crop monitoring, infrastructure inspection, and navigation in confined or elevated spaces.
However, ensuring safe operation in cluttered environments remains a significant challenge.
Unlike ground robots, even minor collisions can result in catastrophic failure for aerial vehicles.
Achieving safety and reliable operation therefore requires accurately modeling the robot’s occupied volume, representing complex scene geometry, and planning motions that guarantee separation between the two.

Existing collision avoidance methods for aerial vehicles typically approximate the robot as spherical or convex \cite{chen2024catnips, chen2025splatnav, kousik2019dronertd, shao2021reachability, richter2016polynomial, tordesillas2020faster, freire2023flatnessbased, ren2022bubble}.
This conservative approximation simplifies computation, but impedes the ability of the robot to operate in cluttered environments.
Another category of planners enforce collision avoidance only at discrete time points using sampling-based planners, inducing a trade-off between safety and computational cost \cite{ragel2015sampling, kim2019sampling, adamkiewicz2021nerfnav, sucan2012ompl}.
This work addresses these limitations by exploiting quadrotors' differential flatness to construct a continuous-time safety representation that tightly overapproximates a robot's swept volume.

Accurate scene representation presents an additional challenge.
Many planning methods assume access to ground-truth scene geometry, which is unavailable in many real-world deployments.
To solve this problem, robotics researchers have increasingly turned their attention to radiance field models such as NeRFs \cite{mildenhall2020nerf} and 3D Gaussian Splats \cite{kerbl20233d} due to their ability to accurately reconstruct complex scene geometry directly from camera data.
However, existing radiance field planners for quadrotors rely on spherical robot approximations \cite{chen2024catnips, chen2025splatnav}.
As illustrated in this paper, this spherical overapproximation limits quadrotors' ability to navigate highly cluttered scenes.
This work addresses these challenges by introducing an optimized continuous-time collision-checking method for quadrotors and scenes represented as normalized 3D Gaussian Splats.

In summary, the contributions of this work are as follows:
\begin{enumerate}
\item A differential flatness-based reachability formulation that constructs a continuous-time forward reachable set for quadrotors, accounting for the full geometry of the robot.
\item  An optimized collision checking method between the forward reachable set and scenes represented as Normalized 3D Gaussian Splats.
\end{enumerate}
The two contributions above are implemented in \algname, a receding-horizon trajectory optimization framework that integrates them to synthesize safe motions in complex environments. Experiments demonstrate state-of-the-art collision avoidance performance for quadrotors across diverse cluttered environments.

\begin{figure}
    \centering
    \includegraphics[width=\linewidth]{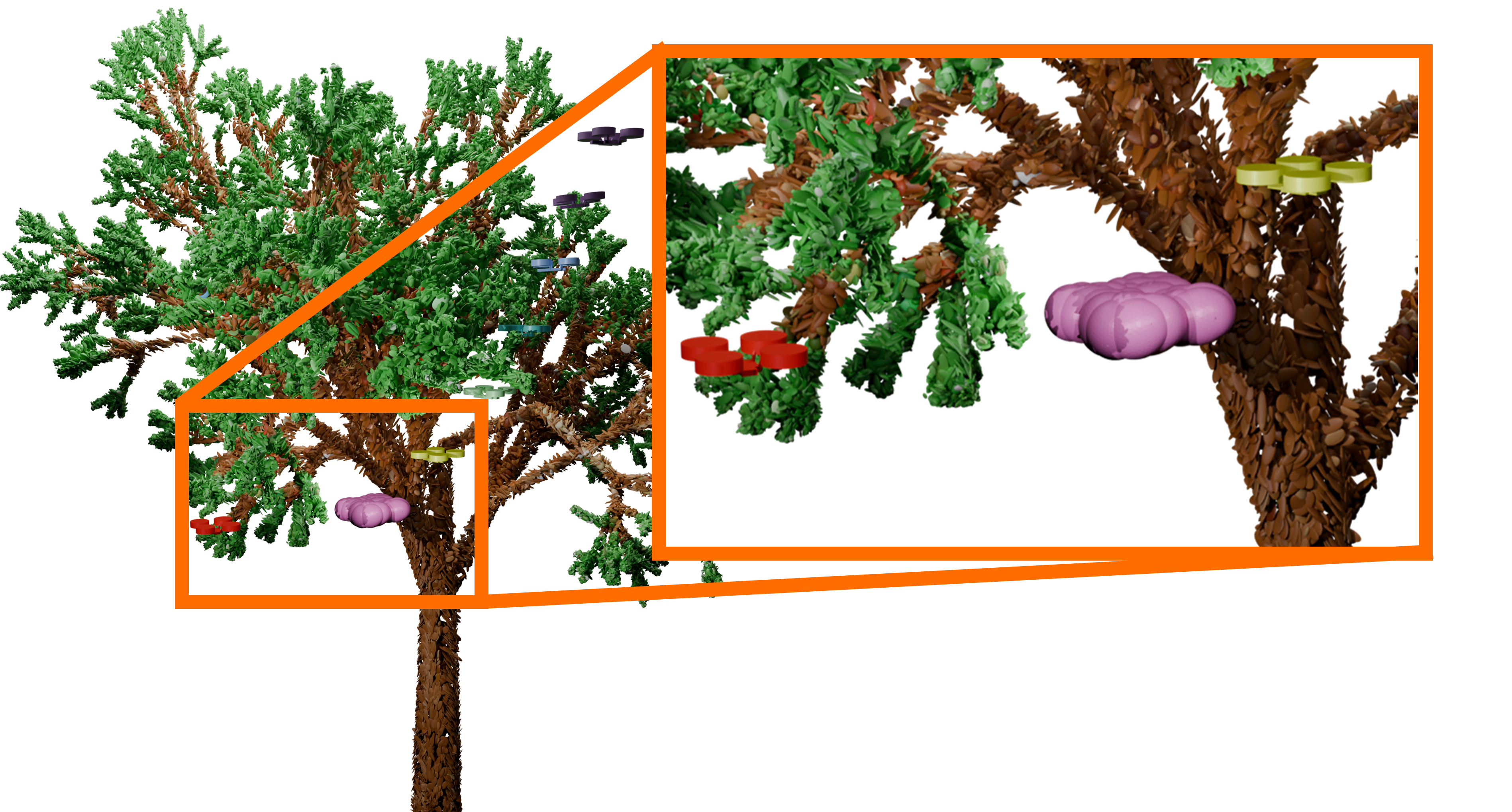}
\caption{\algname generates safe quadrotor trajectories in cluttered environments. The robot occupancy is tightly overapproximated using a reachable set (shown in pink) based on differential flatness. The scene is modeled as a Normalized 3D Gaussian Splat. A sampling-based planner jointly reasons over both representations to compute safe paths between waypoints (yellow) to the goal (red).}
\label{fig:fig1}
\end{figure}

\begin{figure*}[h!]
    \centering
    \includegraphics[width=0.93\linewidth]{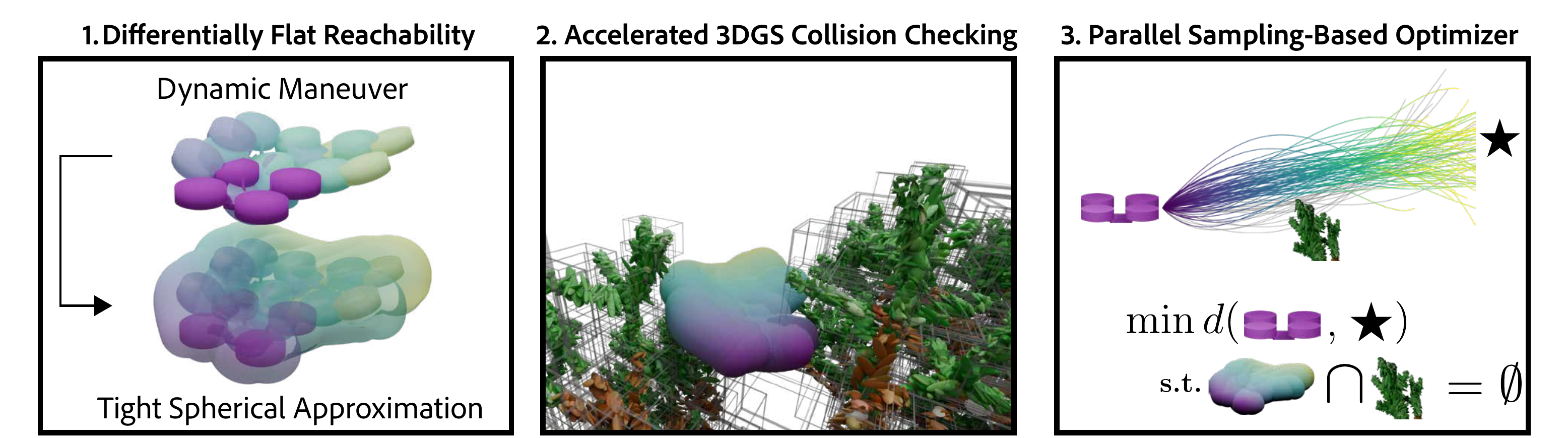}
    \caption{\algname plans safe quadrotor trajectories through cluttered environments represented as normalized Gaussian Splats. It combines three key components.
    First, differential flatness is used to construct a tight sphere-based overapproximation of the robot volume along a trajectory.
    Second, the collision probability between each sphere and the normalized 3D Gaussian Splat is efficiently computed using a Bounding Volume Hierarchy (visualized with gray wireframe boxes).
    Third, a GPU-parallel sampling-based optimizer evaluates many candidate trajectories to efficiently find safe paths through dense environments.}
    \label{fig:method}
\end{figure*}

\section{Related Works}
This paper describes a method that combines collision avoidance for quadrotors, planning in radiance field representations, and reachability-based motion planning. The relevant literatures are summarized below.

\subsection{Quadrotor Collision Avoidance}\label{subsec:rel_drone_collision}

Quadrotors and other aerial vehicles must remain strictly collision-free during operation; yet generating safe plans through cluttered environments remains challenging.
A prominent category of works constructs a global map of geometric primitives and optimizes collision-free plans within it \cite{richter2016polynomial, tordesillas2020faster, kousik2017rtd, freire2023flatnessbased}.
These methods require explicit representations of scene occupancy, such as voxel grids or octrees \cite{tordesillas2020faster, richter2016polynomial}, zonotopes \cite{kousik2019dronertd}, or polytopes \cite{freire2023flatnessbased}.
Polytopes and zonotopes admit efficient constraint formulations but are difficult to construct from sensor data.
Voxel grids and octrees avoid this difficulty but carry high memory overhead in dense, cluttered scenes.
A second category operates reactively to local obstacles without maintaining a global map \cite{lin2025collisionfree, bucki2020rappids}.
While effective with local sensing data, reactive planning makes safety guarantees difficult to establish.
This paper addresses both limitations by performing risk-aware planning directly in high-fidelity radiance fields, which can be reconstructed from sensor data while supporting rigorous safety guarantees.

A recurring pattern in existing methods is overapproximating the robot with a simple primitive such as a sphere \cite{richter2016polynomial, tordesillas2020faster, freire2023flatnessbased, Tayaletal2024, Goswamietal2023}, superellipsoid \cite{lin2025collisionfree}, or zonotope \cite{kousik2019dronertd}. While computationally efficient, this simplification limits safe planning for vehicles with complex geometry in tight spaces. A common workaround is to employ sampling: either checking for collisions at poses sampled along the path \cite{sucan2012ompl}, or sampling points on the robot's surface \cite{adamkiewicz2021nerfnav}.
Sampling, however, makes collision-avoidance guarantees difficult to establish.
In contrast, this paper leverages the differential flatness of quadrotors to compute forward reachable sets, yielding tight, continuous-time overapproximations of the robot's swept volume.

\subsection{Planning in Radiance Fields}
While many safe motion planners assume ground-truth knowledge of obstacle locations, such information is often not available.
Several works have proposed planning methods that operate directly on NeRF reconstructions.
NeRF-Nav \cite{adamkiewicz2021nerfnav} plans dynamically-feasible trajectories for quadrotors by sampling a finite set of points from the robot body, and applying a penalty to the density integrated along those points' paths.
While NeRF-Nav demonstrates the promise of radiance field planning, this approach relies on a discrete approximation of robot geometry and does not provide probabilistic safety guarantees.
CATNIPS overcomes these limitations by relating NeRFs to a Poisson Point Process, and using this relation to convert the NeRF into an occupancy grid~\cite{chen2024catnips}.

While NeRFs use neural networks to approximate radiance fields, 3D Gaussian Splatting represents the same information using a set of unnormalized 3D Gaussians~\cite{kerbl20233d}.
Splat-Nav plans in 3D Gaussian Splats by building safe flight corridors that avoid the $1\sigma$ level sets of these Gaussians \cite{chen2025splatnav}.
Other planning methods avoid collisions with the splats by sampling trajectory points \cite{tao2025rtguide} or applying a soft collision penalty \cite{jin2024gsplanner}.
These offer useful heuristics but assume a spherical robot and provide no continuous-time collision-avoidance guarantees.
Further, it remains unclear how to extend existing methods to robots that are not tightly approximated as spheres.

In contrast to these approaches, Splanning uses a normalized formulation of 3D Gaussian Splatting and applies reachability analysis to overapproximate the continuous-time swept volume of a robot manipulator \cite{michauxisaacson2024splanning}.
It then solves a trajectory optimization that bounds the probability of intersection between the robot and the scene.
Still, Splanning relies on a manipulator fixed to a workspace.
To address these limitations, this paper develops improved reachability, collision-checking, and trajectory optimization methods that enable quadrotors to navigate highly cluttered environments.

\subsection{Reachability-Based Motion Planning}
Reachability-based Trajectory Design (RTD) generates safe, receding-horizon trajectories by constructing a continuous-time representation of the robot's swept volume along candidate trajectories \cite{kousik2017rtd}.
This swept volume, called a Forward Reachable Set (FRS), is checked for collisions with obstacles to provably guarantee safety.
RTD has been applied to ground vehicles \cite{kousik2017rtd}, quadrotors \cite{kousik2019dronertd}, and manipulators \cite{holmes2020armtd, michaux2024sparrows}.
It has also been extended to risk-aware planning with uncertain obstacles \cite{liu2023radius}, and to planning in radiance fields represented by 3D Gaussian Splatting~\cite{michauxisaacson2024splanning}.
Finally, Kwon et al. replace explicit FRS construction with a neural network approximation and use conformal prediction to guarantee conservativeness \cite{kwon2024crows}.
This work extends the literature on reachability-based motion planning by leveraging quadrotor differential flatness to construct a tight safety representation for aerial vehicles, which can be efficiently integrated over a Normalized Gaussian Splat to compute the probability of collision.

\section{Background}\label{background}
This section provides a brief technical background on three components of the proposed method: quadrotor trajectory design, reachability analysis, and the scene representation used for obstacle avoidance.

\subsection{Differential Flatness of Quadrotors}\label{subsec:flatness}
Although quadrotors are underactuated in their full $SE(3)$ state, they are differentially flat \cite{nieuwstadt1996flatness}: the state and control inputs are algebraic functions of four \textit{flat outputs} and their derivatives, namely the position $\pos(t)\in\R^3$ and yaw $\psi(t)\in SO(2)$ \cite{mellinger2011minsnap}.
Any sufficiently smooth curve in $\R^3\times SO(2)$ with bounded derivatives therefore corresponds to a dynamically feasible trajectory, reducing trajectory planning to constructing smooth functions $\pos$ and $\psi$.
The attitude is recovered as follows.
Let $\mathbf{g} = [0, 0, g]^\top$ be the gravity vector in the world frame.
The body-frame $z$-axis aligns with the total thrust direction:
\begin{equation}\label{eq:zB}
\mathbf{z}^B = \frac{\ddot{\pos} + \mathbf{g}}{\|\ddot{\pos} + \mathbf{g}\|}.
\end{equation}
Given $\psi$, the remaining body axes are
\begin{equation}\label{eq:bodyframe}
\mathbf{x}^C = \begin{bmatrix}\cos\psi \\ \sin\psi \\ 0\end{bmatrix},\quad
\mathbf{y}^B = \frac{\mathbf{z}^B \times \mathbf{x}^C}{\|\mathbf{z}^B \times \mathbf{x}^C\|},\quad
\mathbf{x}^B = \mathbf{y}^B\times \mathbf{z}^B,
\end{equation}
where $C$ denotes a frame coincident with the center-of-mass frame $B$ but with zero roll and pitch.
These relations uniquely determine the body rotation $R^{B} = [\mathbf{x}^B \;\; \mathbf{y}^B \;\; \mathbf{z}^B]$: roll and pitch follow entirely from $\ddot{\pos}$ and $\psi$.
Planning can thus occur in $\R^3\times SO(2)$, where the quadrotor is fully actuated, while uniquely recovering the full $SE(3)$ trajectory.

\subsection{Polynomial Zonotopes for Reachability Analysis}

Reachability analysis computes an overapproximation of the states a system may attain over a time interval.
In configuration space, this yields a joint reachable set (JRS), which forward kinematics propagates into the FRS, a conservative overapproximation of the robot's swept volume.
Assuming precise trajectory tracking, this yields a guaranteed bound on swept occupancy for collision checking.

Polynomial zonotopes (PZs) are a set representation that captures nonlinear dependence between variables \cite{kochdumper2020polyzono, michaux2023armour, michaux2024sparrows}.
They support exact propagation through addition and multiplication, corresponding to Minkowski sums and set products.
General nonlinear functions acting on a polynomial zonotope can be efficiently overapproximated, producing a new polynomial zonotope that contains the true image.
In particular, for an analytical function $f: \mathbb{R}^N \to \mathbb{R}^M$, it is possible to evaluate $f$ on a PZ $\mathbf{S} \subset \mathbb{R}^N$ to obtain a PZ $\mathbf{Q} \subset \mathbb{R}^M$ such that $\{f(\mathbf{x}) \mid \mathbf{x} \in \mathbf{S}\} \subseteq \mathbf{Q}$, where we write $f(\mathbf{S}) = \mathbf{Q}$ by an abuse of notation.
For brevity, the mathematical formulation of PZs is omitted; we refer readers instead to \cite{kochdumper2020polyzono, michaux2024sparrows} for a comprehensive treatment.

\subsection{Scene Representation}
Following \cite{michauxisaacson2024splanning}, \algname represents scenes using a normalized 3D Gaussian Splat, a form of radiance field representation that allows probabilistic interpretation of rigid body collisions.
The properties of this representation are summarized below.

\subsubsection{Normalized 3D Gaussian Splatting}
A radiance field is a function $\mathcal{L}:(\mathbf{x},\mathbf{d})\mapsto(r,g,b,\sigma)$ mapping position $\mathbf{x}\in\R^3$ and direction $\mathbf{d}\in\mathbb{S}^2$ to color and density $\sigma\in\R^+$.
Radiance fields can be learned from images via differentiable rendering.
For collision avoidance, color is irrelevant; only $\sigma$ is needed. Because it is independent of $\mathbf{d}$, we write $\sigma(\mathbf{x})$.

In a Normalized 3D Gaussian Splat, $\sigma$ is represented as
\begin{align}
  \sigma(\mathbf{x})=\sum_{n=1}^N w_n G_n(\mathbf{x}),
\end{align}
where $w_n\in\R^+$ and $G_n:\R^3\to\R$ is a Gaussian with mean $\boldmu_n$ and covariance $\Sigma_n$:
\begin{equation}
  G_n(\mathbf{x})=\frac{1}{\sqrt{(2\pi)^3\det{\Sigma_n}}}
  \exp\!\left(-\frac{1}{2}(\mathbf{x}-\boldmu_n)^T\Sigma_n^{-1}(\mathbf{x}-\boldmu_n)\right).
\end{equation}

\subsubsection{Collision Probability}

One can bound the probability that a sphere collides with a 3D Gaussian Splat:

\begin{thm} \cite[Theorem 6]{michauxisaacson2024splanning}\label{thm:collision_bound}
Consider, without loss of generality, a ball $S~=~\Btwo(0,~\rho)$ of radius $\rho$ centered at the origin.
Let $\beta \in [0,1]$ denote a user-set risk threshold.
Then, the probability $\pr(C(S))$ that the ball $S$ collides with the environment is bounded above by
\begin{equation}\label{eq:risk_bound}
    \pr(C(S)) \leq \frac{1}{\beta}\hspace{-2pt}\left[1 - \exp\hspace{-2pt}\left(- \frac{1}{4\pi} \bound(S) \right)\right]
\end{equation}
where
\begin{align}
\bound&(S) = \sum_{n=1}^{n_G} \eta'_n w_n  \cdot \nonumber \\
& \prod_{\ell=1}^3\left[\sqrt{\frac{\pi \lambda'_{n,\ell}}{2}} \left( \erf\left(\frac{\rho - \mu'_{n,\ell}}{\sqrt{2\lambda'_{n,\ell}}}\right)
    -\erf\left(\frac{\rho + \mu'_{n,\ell}}{\sqrt{2\lambda'_{n,\ell}}}\right) \right)\right].
\end{align}
Above, $\boldmu'_n$, $\lambda'_n$, and $\eta'_n$ denote the mean, eigenvalues, and normalization constant of the Gaussian $G'_n$ obtained by rotating $G_n$ by $R_n^T$, i.e., $\boldmu'_n=R_n^T\boldmu_n$ and $\Sigma'_n=R_n^T\Sigma_nR_n$.
\end{thm}
\noindent This bound provides a computationally efficient manner to bound the probability of collision between a rigid body and the scene.

\section{Method}

This section introduces \algname, a method for risk-aware trajectory generation in scenes reconstructed from camera data.

\subsection{System Overview}

\algname synthesizes motion plans by solving the following optimization problem in a receding-horizon manner:
\begin{align}
    \label{eq:optcost}
    &\underset{k\in K}{\min} &&\texttt{cost}(k) \\
    &&& \pr\left(\text{FRS}(q(t; k, x_0)) \cap \mathscr{E} \neq \emptyset\right) \leq \beta   &\forall t \in T \label{eq:optcollision}
\end{align}
A trajectory parameter $k$ is chosen from a parameter space $K$, the cost in \eqref{eq:optcost} encourages the robot to reach a specified goal configuration, and \eqref{eq:optcollision} encodes a risk-aware obstacle avoidance constraint, enforcing that the probability of the robot's FRS intersecting the environment $\mathscr{E}$ may not exceed a risk threshold $\beta$.
Here $x_0$ denotes the robot's initial state (e.g., position, velocity, and acceleration), which is fixed at planning time and not optimized over.
Section~\ref{subsec:reachability} describes the trajectory parameterization and construction of an FRS for an aerial vehicle, while Section~\ref{subsec:collcheck} describes an optimized form of the collision check in Theorem \ref{thm:collision_bound}, and Section~\ref{subsec:trajopt} describes a GPU-parallelized trajectory optimization.

\subsection{Trajectory Parameterization and Reachability Analysis}\label{subsec:reachability}

We compute an FRS for the quadrotor as a union of spheres in three steps: (1) constructing reachable sets of the flat outputs, (2) propagating them to the full system state using differential flatness, and (3)  approximating the resulting FRS with a neural network for efficient online evaluation.

\subsubsection{Trajectory Parameterization and JRS Construction}
Each flat output is parameterized as a degree-5 Bernstein polynomial in time, whose coefficients depend on both the initial state $x_{0}$ and trajectory parameter $k_d$:
\begin{equation}\label{eq:bernstein}
q_d(t;\, x_{0}, k_d) = \sum_{l=0}^{5} \mathcal{B}_{d,l}(x_{0}, k_d)\, b_l(t),
\end{equation}
where $b_l$ are the Bernstein Basis Functions \cite[Example 15]{michaux2023armour}.

To ensure safe receding-horizon planning, the planning horizon $T$ is divided into two intervals: the first corresponds to the planned motion, and the second to a braking maneuver that brings the system to zero velocity and acceleration.
During execution of the first interval, the system plans the next horizon; if successful, the braking maneuver is discarded and planning continues in a receding-horizon manner.
For convenience, we set $ T=1$ second and divide it equally between the two intervals.

Because the robot executes the previously planned trajectory while replanning, $x_{0}$ is known at planning time.
Furthermore, since every trajectory must end in a braking maneuver, the terminal velocity and acceleration are fully determined.
Together, these boundary conditions fix five of the six Bernstein coefficients.
We parameterize the remaining coefficient, which we choose without loss of generality to be the last coefficient in \eqref{eq:bernstein}, as
\begin{equation}
    \mathcal{B}_{d,5}(k_d) = \eta_{d,1} k_d + \eta_{d,2},
\end{equation}
where $\eta_{d,1}, \eta_{d,2} \in \mathbb{R}$ are user-specified constants.
As a result, $q_d(t;\, x_{0}, k_d)$ is polynomial in $k_d$ and $t$.
By representing $[0,T]$ as a PZ and evaluating \eqref{eq:bernstein}, we obtain a PZ $\mathbf{Q}_d$ that bounds all reachable configurations of DOF $d$ over the planning horizon as a function of $k_d$.
We refer to $\mathbf{Q}_d$ as the $JRS$ of the $d^{th}$ DOF of the system.
In practice, the JRS of the quadrotor system is built by assembling four of the individual $\mathbf{Q}_d$ objects.
Each position dimension is represented by a single JRS axis.
Yaw is represented using separate JRSs for $\sin(\psi)$ and $\cos(\psi)$ to reduce nonlinear overapproximation error, as is common in reachability-based methods \cite{michaux2024sparrows}.
We refer to the 5-DOF JRS of the quadrotor's position and yaw as $\mathbf{Q}$.

\subsubsection{Computing the SE(3) FRS}
While $\mathbf{Q}$ bounds the set of states that the robot's position and yaw will occupy during a trajectory, there remains ambiguity in the roll and pitch angles.
While prior works commonly avoid this ambiguity by overapproximating the robot with a sphere \cite{chen2025splatnav, chen2024catnips} or zonotope \cite{kousik2019dronertd}, we instead propagate the JRS through differential flatness relations \eqref{eq:zB}--\eqref{eq:bodyframe}, yielding a reachable rigid-body transform PZ $\mathbf{T}^{WB}$ .

To obtain an FRS that bounds the quadrotor conservatively, we pack the robot body with body-fixed spheres with centers $\{\mathbf{c}_m\}$ and radii $\{r_m\}$ generated using the method of \cite{coumar2025foam}.
Each center is propagated through the set of reachable rigid body transformations:
\begin{equation}
    \mathbf{C}_{m} = \mathbf{T}^{WB} \mathbf{c}_m.
\end{equation}
This produces a reachable set for each sphere center.
Each set $\mathbf{C}_{m}$ is then overapproximated by a bounding sphere with center $\mathbf{c}_{m}$ and radius $\delta_{m}$.
The corresponding robot volume is obtained by inflating this radius by the body-fixed sphere radius, yielding a final radius
\begin{equation}
    R_{m} = \delta_{m} + r_m.
\end{equation}
Finally, the robot FRS over the time interval is given by the union of the spheres $B(\mathbf{c}_{m}, R_{m})$ over all $m$.
In practice, the planning horizon is split into several sub-intervals, and the spheres from all sub-intervals are accumulated to form the full FRS.
The resulting spheres bound the swept volume of the aerial base.
The overapproximativeness of this bound follows closely from \cite[Theorem 10]{michaux2024sparrows}, where we apply the differential flatness relations instead of the forward kinematics of arms.

\subsubsection{Neural Network Approximation}

The computation above yields an overapproximation of the robot's occupancy. In practice, however, PZs come with a trade-off between the tightness of the overapproximation and computational complexity.
The compute requirements of highly-accurate PZs limit applicability to the sampling-based planner we describe in Section \ref{subsec:trajopt}.
To reduce online computation, we adopt the method of \cite{kwon2024crows} and train a neural network to approximate the spherical FRS.
With these values, the corresponding spherical FRS is computed using the exact reachability pipeline.
The network predicts sphere centers and radii as functions of these inputs, and conformal prediction is applied to conservatively inflate the outputs and preserve safety guarantees.

\subsection{Parallel Collision Detection}\label{subsec:collcheck}

In practice, each FRS contains hundreds to thousands of spheres per planning horizon.
The sampling-based planner described in Section \ref{subsec:trajopt} may consider up to 100 candidate trajectories, requiring collision checking on up to hundreds of thousands of spheres per iteration.
Further, the environment representation may contain millions of Gaussians.
Naively evaluating the collision constraint in \eqref{eq:risk_bound} across all sphere-Gaussian pairs is therefore computationally prohibitive.
To overcome this problem, Splanning \cite{michauxisaacson2024splanning}  ignores Gaussians outside the reachable workspace of the fixed manipulator.
Yet, this solution does not handle a mobile vehicle and conservatively assumes that all spheres interact with the same set of Gaussians.
Instead, we use a Bounding Volume Hierarchy (BVH) to accelerate collision queries by restricting evaluation to nearby Gaussians \cite{karras2012bvh}.

\subsubsection{BVH construction}

For each Gaussian, an axis-aligned bounding box (AABB) is computed corresponding to the Gaussian's $n_\sigma$ confidence region.
A linear BVH is then constructed over these AABBs using an open-source implementation\footnote{https://github.com/ToruNiina/lbvh} of the method in \cite{karras2012bvh}.
This pre-processing happens once per scene.

\subsubsection{Query procedure}

Each sphere is processed by a GPU thread.
During a query, the sphere's AABB is computed, then the BVH traversed to identify overlapping Gaussians.
Each thread stores candidate Gaussians in a per-thread buffer of size $n_B$, capping the number of Gaussians considered for that sphere.
Let $n_{K, i} \le n_B$ denote the number of Gaussians retrieved for sphere $i$.
The collision risk is computed by evaluating the closed-form integral from Theorem~\ref{thm:collision_bound} over this candidate set.
If the buffer capacity is exceeded during traversal (i.e., more than $n_B$ Gaussians overlap the sphere), the trajectory candidate is immediately flagged as unsafe.

\subsubsection{Parallelization and Complexity}

The BVH reduces the number of Gaussian evaluations per sphere from $n_G$ to $n_{K, i} \ll n_G$, yielding $\mathcal{O}(\log n_G + n_{K,i})$ time complexity and $\mathcal{O}(n_B)$ memory complexity for each sphere query.

\begin{algorithm}[t]
\caption{Sampling-Based Trajectory Optimization}
\label{alg:softcem_trajopt}
\begin{algorithmic}[1]
\Require Mean $\mu_0$, covariance $\Sigma$, temperature $\lambda$, samples $n_S$, iterations $n_I$, max time $T_{\max}$, collision weight $w_{\mathrm{col}}$, tolerance $\beta$
\Ensure Trajectory parameter $k^*$, or $\varnothing$ if none is safe
\State $k^* \gets \varnothing,\ c^* \gets \infty$
\For{$i = 0, \dots, n_I-1$}
  \If{elapsed time $> T_{\max}$} \textbf{break} \EndIf
  \For{$j = 1, \dots, n_S$ \textbf{(in parallel)}}
    \State $k_j \sim \mathcal{N}(\mu_i, \Sigma)$, clipped to $[-1,1]$
    \State $\texttt{cost}_j \gets \texttt{cost}(k_j) + w_{\mathrm{col}}\, c_{\mathrm{col}}(k_j)$
  \EndFor
    \State $j' \gets \arg\min_j \texttt{cost}_j$
    \If{$\texttt{cost}_{j'} < c^*$}
      \State $k^* \gets k_{j'},\ c^* \gets \texttt{cost}_{j'}$
    \EndIf
    \State $w_j \gets \exp(-\texttt{cost}_j/\lambda) \big/ \sum_{\ell} \exp(-\texttt{cost}_\ell/\lambda)$
  \State $\mu_{i+1} \gets \sum_{\ell} w_\ell \, k_\ell$
\EndFor
\If{$k^* = \varnothing$ \textbf{ or } $c_{\mathrm{col}}(k^*) > \beta$}
  \State \Return $\varnothing$
\EndIf
\State \Return $k^*$
\end{algorithmic}
\end{algorithm}

\subsection{Trajectory Optimization}\label{subsec:trajopt}
Most reachability-based motion planners leverage interior point methods to optimize trajectories subject to collision avoidance constraints \cite{michauxisaacson2024splanning}.
In practice, this style of optimization fails to converge on the differentially flat quadrotor system.
We suspect this is because there are spurious local minima in the optimization problem.
Hence, we instead use a sampling-based trajectory optimizer based on a soft cross-entropy method \cite{rubinstein2004cross, williams2018mppi}.
As described in Algorithm \ref{alg:softcem_trajopt}, trajectories are iteratively refined using importance-weighted averaging of Gaussian samples.
The user-specified cost is augmented with a collision penalty computed as described in Section \ref{subsec:collcheck}.
Trajectories are rejected if the final collision probability exceeds the risk threshold.

\section{Experiments}
Two experiments were run to evaluate the proposed methodology.
First, the optimized collision checker was evaluated for speed and accuracy.
Second, the proposed planner was evaluated end-to-end against relevant baselines.
All experiments were run on a machine with an AMD 5950x CPU and NVIDIA RTX Pro 6000 GPU.

\subsection{Collision Constraint}

The collision constraint was evaluated according to the procedure described in \cite{michauxisaacson2024splanning}.
The collision comparison operates on the randomized scenes with cube obstacles.
Individual spheres were randomly sampled that were either in collision, near collision, or collision-free with an equal distribution of each.
Each method then evaluates whether each sphere is in collision with the environment.

\subsubsection{Collision Baselines}
Three baseline methods were evaluated.
First, the collision constraint from CATNIPS between a sphere and a NeRF was run using the reference implementation and the default settings \cite{chen2024catnips}.
Second, collision comparisons between a standard, un-normalized 3DGS and a sphere was evaluated using the constraint in Splat-Nav \cite{chen2025splatnav}.
For Splat-Nav, the constraint was evaluated by checking for collisions between the sphere and multiple $n\sigma$ ellipsoids of the un-normalized Gaussians using the reference implementation.
Next, the Splanning constraint between spheres and a normalized 3D Gaussian Splat \cite{michauxisaacson2024splanning} was evaluated at multiple risk thresholds $\beta$ (See Thm \ref{thm:collision_bound} for a definition of $\beta$).
Finally, the optimized Splanning constraint proposed in this method was evaluated.
Note that while the constraint in \cite{michauxisaacson2024splanning} and that proposed in this paper are theoretically identical, differences in the logic for ignoring far-away Gaussians and numerical implementation may cause slight discrepancies.

\subsubsection{Metrics}
Each collision checker was treated as a classifier of free space vs collision, and both precision and recall were evaluated at a range of parameters.
Furthermore, for the Splanning constraint and our method, the time required to evaluate the constraint was measured.
For Splat-Nav and CATNIPS, the compute time was not measured because the methodology required to isolate the collision constraint from the rest of the planner significantly affected their runtime performance.

\subsection{Motion Planning}
\begin{figure}
    \centering
    \includegraphics[width=\linewidth]{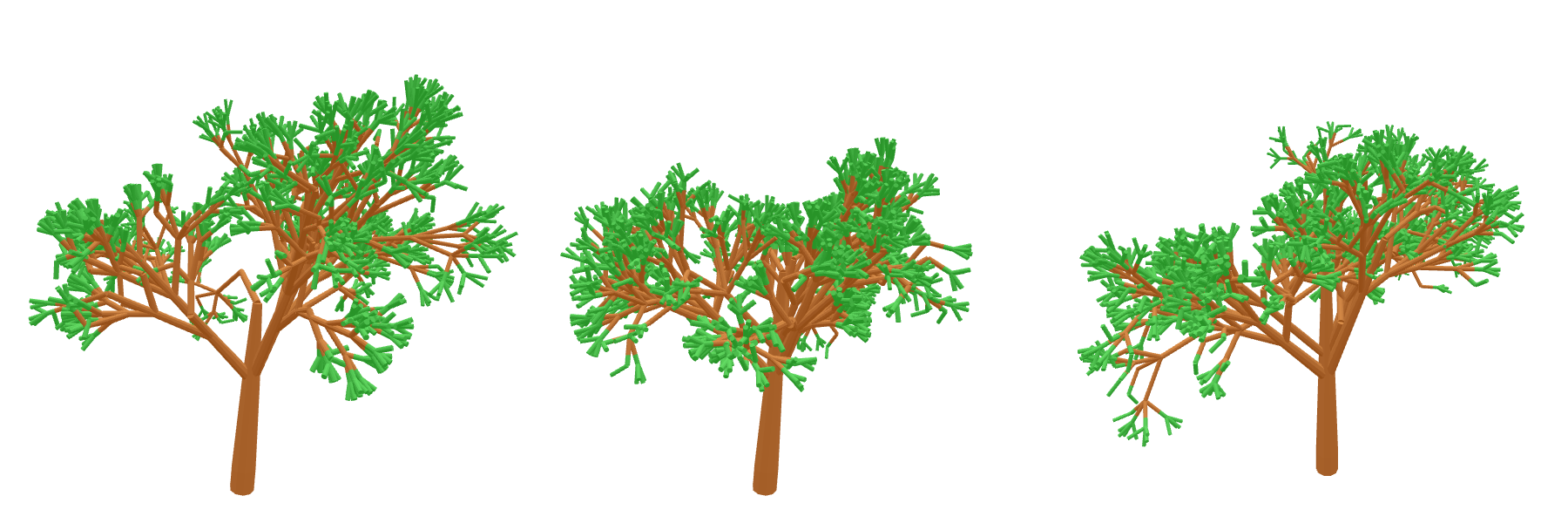}
    \caption{Examples of images used to reconstruct the tree environments used in planning experiments. Trees were procedurally generated using L-Systems. 3DGS models were reconstructed from these simulated observations.}
    \label{fig:training_images}
\end{figure}

To create realistic, challenging obstacle-avoidance environments, we procedurally generate three-dimensional trees using L-systems, representing each segment as a ten-sided zonotope.
Scenes are rendered in PyRender with cameras arranged spherically around each tree, emulating images captured by an aerial vehicle at a safe distance (Figure~\ref{fig:training_images}).
These images are used to train a normalized 3D Gaussian Splat for \algname, a standard 3D Gaussian Splat \cite{kerbl20233d} for Splat-Nav \cite{chen2025splatnav}, and a NeRF \cite{mildenhall2020nerf} for CATNIPS \cite{chen2024catnips}.
Only our method and CATNIPS used depth supervision, as Splat-Nav's Nerfstudio pipeline does not support it for 3D Gaussian Splatting \cite{nerfstudio}.
We generate 100 unique tree scenes, each with five pairs of collision-free start and goal configurations for a simulated quadrotor, though a collision-free path between them is not guaranteed to exist.

The simulator verifies successfully generated trajectories of each method by checking for intersections between the robot and the ground truth obstacle meshes.
In the case of \algname, the simulator checks for collisions after each time horizon, while for \cite{chen2025splatnav} and \cite{chen2024catnips}, collision comparisons are performed post-hoc using the optimized trajectories.
In these experiments, \algname is run with high-level waypoints computed with a Bidirectional RRT \cite{sucan2012ompl} where constraint evaluation is implemented using the same constraint as defined in Theorem~\ref{thm:collision_bound}.
The optimizer defined in Algorithm~\ref{alg:softcem_trajopt} is run in a receding-horizon fashion and the optimizer is constrained to 0.45s to compute each second of motion; if this optimizer fails, the safe braking maneuver is executed, and the planner continues. If after ten planning horizons the vehicle has not moved more than 10cm, the RRT is re-run from the current state. Each trial is run for a maximum of 150 planning horizons.

A success indicates that the quadrotor reached the goal with zero collisions. Failures are counted if the robot fails to reach the goal or if the planned path results in a collision.
The parameters used in the motion planning experiments are listed in Table \ref{tab:params}.

\begin{table}[]
    \centering
    \caption{Key planning, reachability, and collision-checking parameters used in the motion planning experiments}
    \begin{tabular}{l|l|c}
    \hline
    Symbol & Description & Value \\
    \hline
    $n_\sigma$ & Gaussian culling level & 10 \\
    $n_B$ & BVH buffer size & 16,384 \\
    $\beta$ & Risk threshold & 0.01 \\
    $T$ & Planning horizon & 1s \\
    $n_S$ & Number of trajectory samples & 96 \\
    $n_I$ & Max optimizer iterations & 20 \\
    $T_{\mathrm{max}}$ & Optimizer Timeout & 0.45s \\
    $\lambda$ & Cross-entropy temperature & 1.0 \\
    $w_{\mathrm{col}}$ & Collision penalty weight & $10^4$ \\
    $n_{\mathrm{sph}}$ & Num. spheres in spherepacking & 25 \\
    \hline
    \end{tabular}
    \label{tab:params}
\end{table}

\subsection{Ablation Study}\label{subsec:ablations}
Two key features of the system were individually assessed for their contribution to the overall system.
First, the flatness-based FRS was replaced with a simple single-sphere overapproximation, mirroring a common approach in the literature \cite{chen2024catnips, chen2025splatnav, richter2016polynomial, tordesillas2020faster, freire2023flatnessbased}.
Second, the PZ reachability pipeline proposed in this paper was replaced with that of~\cite{kousik2019dronertd}.
However, unlike \cite{kousik2019dronertd} which over-approximates the robot with a single zonotope, we instead use a sphere to maintain compatibility with the 3DGS collision check.

\section{Results}

This section summarizes the results of our experimental evaluations.

\begin{figure}
    \centering
    \includegraphics[width=0.9\linewidth]{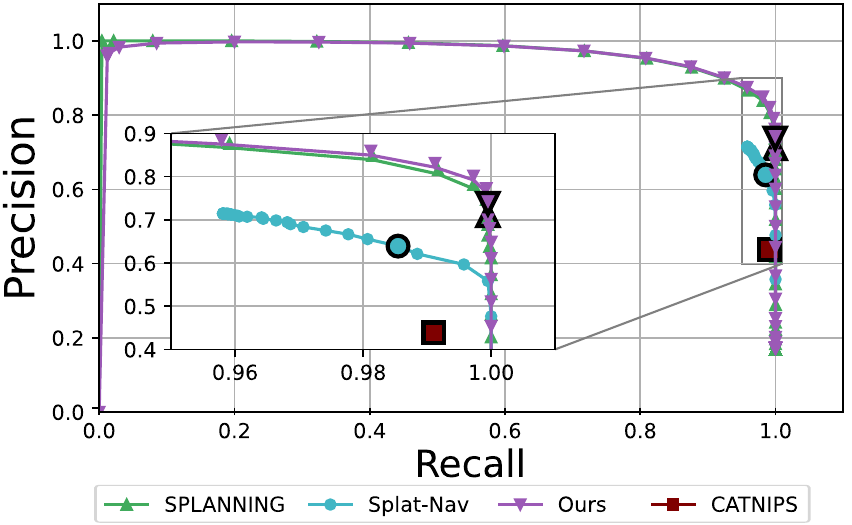}
    \caption{The precision and recall of various collision-checking methods between spheres and radiance fields are evaluated with multiple risk parameters; outlined entries indicate the values used by each method during planning. The proposed collision check yields results that are equivalent to or better than the baselines, with significantly lower computational complexity.}
    \label{fig:prec_rec}
\end{figure}

\subsection{Collision Constraint}

Figure~\ref{fig:prec_rec} shows the precision-recall curves computed for each collision constraint.
CATNIPS achieves high recall but lower precision than other baselines, indicating that it will result in safe but conservative trajectories.
Splat-Nav provides a reasonable balance between precision and recall; however, the comparatively low precision suggests overly conservative behavior.
Splanning and our method produce nearly identical results, as is expected due to their shared probabilistic formulation.
In a finding consistent with that of Splanning \cite{michauxisaacson2024splanning}, the collision bound of Theorem~\ref{thm:collision_bound} provides more accurate collision detection than other baselines.

Benchmarking results are found in Table \ref{tab:compute}.
The BVH-accelerated collision check is, on average, nearly three times faster than the naive PyTorch implementation, with significantly lower standard deviation.
We believe that most of this improvement comes from avoiding the materialization of large intermediate matrices required by the PyTorch implementation.

\begin{table}
  \centering
  \caption{Mean and standard deviation of time required to evaluate the collision constraint.}
  \label{tab:compute}
  \begin{tabular}{l|cc}
    \hline
    Method & Mean Time & Std. Dev.  \\
    \hline
    Splanning & 122.1 ms & 89.1 ms \\
    Ours &\textbf{ 42.2 ms}  & \textbf{4.9 ms} \\
    \hline
  \end{tabular}
\end{table}

\subsection{Planning}
Table~\ref{tab:planning_results} summarizes the end-to-end planning results across 100 procedurally generated tree scenes with 5 start and goal configurations per scene.
In addition to success, stuck, and collision rates, we further report the success weighted by path length (SPL) \cite{anderson2018spl}.
Splat-Nav achieves a success rate of 51.2\%, and becomes stuck on the remaining 48.8\% of trials.
The `stuck' trials resulted from Splat-Nav classifying the start or goal positions as infeasible, indicating the spherical overapproximation of the robot was too conservative for the highly cluttered test scenes.
CATNIPS similarly becomes stuck in 70.4\% of the scenes, which is consistent with the finding in Figure~\ref{fig:prec_rec} that the CATNIPS constraint has high recall but comparatively low precision.
In contrast, \algname succeeds in 81.2\% of scenes without collisions, demonstrating its ability to safely navigate cluttered scenes.

\begin{table}[]
    \caption{Each planner ran 500 trials total, then collision-checked against the scene. SPL is success weighted by path length~\cite{anderson2018spl}.}
    \centering
    \begin{tabular}{l||c|c|c|c}
    \hline
    Method & \% Success & \% Stuck & \% Collide   & SPL \\
    \hline\hline
    Splat-Nav & 51.2\% & 48.8\% & \textbf{0}\% & 0.47 \\
    CATNIPS & 28.8\% & 70.4\% & 0.8\% & 0.26 \\
    \hline
    Ours & \textbf{81.2\%} &\textbf{18.8\%} & \textbf{0\%} & \textbf{0.66}\\
    \hline
    \end{tabular}
    \label{tab:planning_results}
\end{table}
\subsection{Ablation Study}
Results of the ablation study are presented in Table~\ref{tab:system-ablation}.
In the first row, we replace our proposed PZ reachability pipeline with the zonotope reachability method of \cite{kousik2019dronertd}.
This produces a significantly more conservative planner, with success dropping to 16.2\% and the robot becoming stuck in 83.8\% of trials.

Enabling PZ reachability and the sampling-based optimizer raises the success rate to 49.8\%.
This indicates that polynomial zonotopes provide a less conservative safety representation, even without the flatness-based FRS, but still do not overcome the conservativeness of a single-sphere robot approximation.
Enabling flatness-based reachability with a tight sphere packing of the robot but disabling the sampling optimizer reduces the success rate to 25.2\%. This reinforces the importance of the sampling-based optimizer.
Finally, enabling all proposed features increases the success rate to 81.2\%, outperforming all ablations and baselines.

\begin{table}[t]
    \caption{Ablation study over key system components. Each row replaces proposed components with existing baselines.}
    \centering
    \begin{tabular}{ccc|ccc}
    \hline
    \textbf{Flat} & \textbf{PZ} & \textbf{Samp.} & \textbf{Success} & \textbf{Stuck} & \textbf{Collision}  \\
    \textbf{FRS} & \textbf{Reach.} & \textbf{Opt.} & \textbf{\%} & \textbf{\%} & \textbf{\%}\\
    \hline\hline
    \xmark & \xmark & \xmark & 16.2\% & 83.8\% & \textbf{0\%} \\
    \xmark & \cmark & \cmark & 49.8\% & 50.2\% & \textbf{0\%} \\
    \cmark & \cmark & \xmark & 25.2\% & 74.8\% & \textbf{0\%} \\
    \hline
    \cmark & \cmark & \cmark & \textbf{81.2\%} & \textbf{18.8\%} & \textbf{0\%} \\
    \hline
    \end{tabular}
    \label{tab:system-ablation}
\end{table}

\section{Conclusions and Future Work}
This paper introduced \algname, a novel method that synthesizes risk-aware trajectories for quadrotors in scenes represented using radiance fields.
Three key contributions were presented.
First, this paper introduced a novel reachability method that leverages differential flatness to tightly overapproximate the robot's forward reach set.
Second, a highly parallelizable method was developed for collision checking between spheres and a normalized 3D Gaussian Splat.
Finally, a sampling-based optimizer was presented that leveraged the parallel collision check and tight reachability analysis to synthesize risk-aware plans for quadrotors.

Experiments demonstrate that, unlike state-of-the-art methods, the proposed system plans effectively in cluttered environments. Across 500 planning problems in procedurally generated tree structures, \algname successfully found a path in 81.2\% of cases, a substantial improvement over the strongest baseline, which achieved a 51.2\% success rate.
A key takeaway from both the planning experiments and the ablation study is that overapproximating an aerial vehicle with a single sphere limits its ability to generate plans in highly cluttered environments.
These results suggest that coupling high-fidelity scene representations with continuous-time reachability analysis enables aerial robots to identify safe paths that would otherwise be rejected by conservative geometric approximations.

Despite the promising results, the current work has limitations that should be addressed in future work.
First, as with other safe motion planners operating in radiance fields \cite{michauxisaacson2024splanning, chen2024catnips, chen2025splatnav}, \algname assumes that the robot has access to an accurate radiance field representing the scene.
While these can be constructed from sensor data, future work will focus on synthesizing plans in incomplete maps generated by radiance-field-based SLAM algorithms.
Similarly, current evaluations are limited to simulation.
While prior works demonstrate that 3DGS as a representation transfers well between simulation and the real world \cite{michauxisaacson2024splanning}, future work will validate AirSplan on hardware.

Furthermore, the current method assumes that the robot can accurately track the planned motion.
Methods for incorporating tracking error should be considered in the future to increase robustness in real-world deployments \cite{michaux2023armour, kousik2019dronertd}. Finally, the method's computational cost currently requires offboard planning; mitigating this remains future work.

\bibliographystyle{IEEEtran}
\bibliography{IEEEabrv, references}

\end{document}